# Generalization Bias in Large Language Model Summarization of Scientific Research

Uwe Peters*
Dept. of Philosophy, Utrecht University, Netherlands

Benjamin Chin-Yee
Dept. of Pathology and Laboratory Medicine, and Dept. of Medicine, Western University, Canada
Dept. of History and Philosophy of Science, University of Cambridge, UK

*Corresponding author

**Abstract**

Artificial intelligence chatbots driven by large language models (LLMs) have the potential to increase public science literacy and support scientific research, as they can quickly summarize complex scientific information in accessible terms. However, when summarizing scientific texts, LLMs may omit details that limit the scope of research conclusions, leading to generalizations of results broader than warranted by the original study. We tested 10 prominent LLMs, including ChatGPT-4o, ChatGPT-4.5, DeepSeek, LLaMA 3.3 70B, and Claude 3.7 Sonnet, comparing 4900 LLM-generated summaries to their original scientific texts. Even when explicitly prompted for accuracy, most LLMs produced broader generalizations of scientific results than those in the original texts, with DeepSeek, ChatGPT-4o, and LLaMA 3.3 70B overgeneralizing in 26–73% of cases. In a direct comparison of LLM-generated and human-authored science summaries, LLM summaries were nearly five times more likely to contain broad generalizations (OR = 4.85, 95% CI [3.06, 7.70], $p < 0.001$). Notably, newer models tended to perform worse in generalization accuracy than earlier ones. Our results indicate a strong bias in many widely used LLMs towards overgeneralizing scientific conclusions, posing a significant risk of large-scale misinterpretations of research findings. We highlight potential mitigation strategies, including lowering LLM temperature settings and benchmarking LLMs for generalization accuracy.

## 1. Introduction

Accurately communicating findings of scientific studies is vital for educating the public, informing policy, guiding behaviour, and advancing research [1,2]. To learn about, review, and communicate scientific findings, both experts (e.g. researchers) and laypeople (e.g. reporters and students) now increasingly use artificial intelligence (AI) chatbots (e.g. ChatGPT) powered by large language models (LLMs) [3–5]. AI chatbots can process vast amounts of scientific information and summarize content in easily understandable language, thus helping to spread scientific knowledge, promote evidence uptake, and facilitate research [3,6,7].

However, many experts have voiced concerns, noting that AI chatbots used as science communication tools may generate plausible sounding but false or misleading information [3,8–10]. One important related yet underexplored issue is that chatbots may overlook uncertainties, limitations, and nuances in original research by omitting qualifiers and oversimplifying text [11,12], leading to overgeneralizations, i.e. generalizations that are broader than those in the original text and that may therefore be unwarranted by the original findings.

This can result in widespread misinterpretations of findings, illusions of understanding, research lacunas, and risky practices [13]. For instance, LLM chatbots are increasingly used in medical education and clinical practice for research summarization and answering medical queries [10,14,15]. If chatbots produce summaries that overlook qualifiers or restrictors to the generalizability of clinical trial results, trainees and practitioners who rely on these chatbots may prescribe unsafe or inappropriate treatments.

Several recent studies found that scientists and science reporters also frequently overgeneralized or exaggerated scientific findings in their writings [16–18]. This problem could be exacerbated or mitigated if LLMs, instead of human communicators, convey scientific results. However, the specific question of whether LLMs accurately capture the generalizations of scientific research remains unexamined, leaving a critical knowledge gap regarding the societal risks of using LLMs for science summarization that has led several commentators to call for a systematic investigation [11].

To address this gap, we tested 10 prominent LLMs on their ability to summarize abstracts and articles from top journals in science (e.g. *Science*, *Nature*) and medicine (e.g. *The New England Journal of Medicine*, *Lancet*) (see Methods). The models, tested through an application programming interface (API) or website user interface (UI), were GPT-3.5 Turbo (API and UI), GPT-4 Turbo (API and UI), LLaMA 2 70B (API), Claude 2 (API), ChatGPT-4o (UI), ChatGPT-4.5 (UI), LLaMA 3.3 70B Versatile (API), Claude 3.5 Sonnet (UI), Claude 3.7 Sonnet (UI), and DeepSeek (UI). By 'GPT-3.5 Turbo (UI)' and 'GPT-4 Turbo (UI)', we mean ChatGPT-3.5 and ChatGPT-4, respectively, as these systems were powered by GPT-3.5 Turbo and GPT-4 Turbo at the time of the first data collection.

The first four models were selected because they were among the most widely used LLMs at the time of study inception (January 2024), and prior research found that LLaMA 2 and GPT models outperformed humans in medical text summarization [19], while Claude models demonstrated greater faithfulness in book summaries than GPT [20]. To assess diachronic trends in LLM generalization behaviour, the four older models were compared to the six newer ones (tested in March 2025), which currently rank among the most widely used and preferred by scientists [21].

Our primary focus was on GPT models, as they remain dominant LLMs [22], with ChatGPT usage among US teenagers for schoolwork doubling from 13% in 2023 to 26% in 2025 [23]. Additionally, GPT models have been found to produce a lower percentage of misrepresentations (15%) in news summarization compared to competitors such as Perplexity (17%), Copilot (27%), and Gemini (34%), further justifying our emphasis on them [24]. DeepSeek was included due to its rapid rise in popularity, having recently overtaken ChatGPT as the most downloaded free chatbot app [25].

For the scientific texts to be summarized, abstracts (100 from multidisciplinary science journals and 100 from medical journals) were our primary focus as they provide an efficient format for testing summarization by LLMs [9]. Additionally, we tested several models on their summarization of 100 full-length articles, focusing on articles reporting clinical studies because overly broad generalizations of clinical findings can be particularly problematic, often directly affecting policy-making or patient care [18,26]. To systematically assess differences between LLM-generated and human-written summaries, we also collected the corresponding expert-written summaries from NEJM Journal Watch (henceforth '*NEJM JW*') [27].

In our analysis, we compared the generalizations within the result claims of LLM summaries with the generalizations in the original texts. Furthermore, LLM article summaries were compared with NEJM JW summaries of the same articles. Original texts and summaries were coded based on whether their result claims contained one or more of the following three types of generalizations:

(1) *Generic generalizations (generics).* These are present tense generalizations that do not have a quantifier (e.g. 'many', '75%') in the subject noun phrase and describe study results as if they apply to whole categories of people, things, or abstract concepts (e.g. 'parental warmth is protective') instead of specific or quantified sets of individuals (e.g. study participants) [28]. Generics are known to obscure differences between individuals of a reference class since they are semantically underdetermined (e.g. the generic 'children like sweets' may refer to some, most, or all children) [18,26]. Hence, when an LLM summarizes a quantified generalization by using a generic, it transitions from a narrower to a potentially unwarranted broader generalization.

(2) *Present tense generalizations.* Result claims in past tense have a more limited generalization scope than present tense result claims because they refer to a particular sample and do not extend findings to the here and now [18]. When past tense result claims from an original text are turned into present tense in the summary, a broader generalization is conveyed than the author(s) of the original text may have intended [29].

(3) *Action guiding generalizations.* While result claims commonly manifest in descriptive statements (e.g. 'OCD patients benefit from CBT'), they often underlie recommendations (e.g. for policy-makers, practitioners, etc.) about a particular policy or action (e.g. 'CBT should be recommended for OCD patients') [30]. When descriptive result claims are summarized such that action guiding recommendations are communicated, this involves a broader generalization than that found in the summarized text because researchers may have deliberately avoided such recommendations due to insufficient evidence to support them.

We tested whether the outputs of the 10 LLMs mentioned above retained the quantified, past tense, or descriptive generalizations of the scientific texts that they summarized, or transitioned to unquantified (generic), present tense, or action guiding generalizations. We defined the latter kind of conclusions collectively as *generalized* and the former as *restricted conclusions*.

Using logistic regressions to model the scope of a text's conclusion (generalized vs. restricted) as the binary outcome variable, we examined whether LLM summaries of original texts differed from the original texts in the likelihood of containing generalized conclusions. Moreover, we compared the number of original texts containing generalized conclusions with the number of corresponding LLM summaries containing them. When the latter number was higher than the former, this difference indicated the overall cases in which LLMs deviated in their summaries from original texts by producing broader conclusions than the original texts contained. We defined such a case as an *overall algorithmic overgeneralization*. When a specific original scientific text did not contain a generalized conclusion, but the corresponding LLM summary contained one, this was defined as a *specific algorithmic overgeneralization*.

Not all generic, present tense, or action guiding generalizations – whether made by scientists or LLMs – are problematic. When evidentially warranted, these generalizations (by humans) are an essential part of inductive scientific knowledge acquisition [31] and sometimes necessary for effective science communication, as, for instance, members of the public are interested in what the results mean for them now (versus only the sample tested). Similarly, while generic statements carry semantic risks due to their underdetermined meaning [18], they can also be effective in simplifying complex information, making scientific content more accessible.

However, when generalizations lack sufficient empirical support, for instance, when researchers fail to control for confounders or use unrepresentative samples, they become problematic. In this study, we did not assess whether the generalizations in human-authored texts were warranted. Rather, we used them as a baseline for comparison. The faithful representation of the original text served as the normative standard, and we defined 'overgeneralizations' as cases where LLMs broadened conclusions beyond those presented in the original scientific text. To the extent that an LLM user asks specifically only for a summary of a given text, any deviation in generalization from the original remains an epistemically problematic LLM output.

Prior research found that the content of LLM prompts can significantly affect output accuracy [32,33]. Whether this also applies to the accuracy of LLM generalizations in science text summarization has not yet been studied. We therefore also tested three different prompts. The first one simply asked LLMs to summarize a given text without further instruction. The second was selected based on evidence from a previous study, which found that a prompt with the phrase 'take a deep breath and work on this problem step by step' produced LLM outputs with the highest accuracy compared to prompts with more neutral language [32]. While caution is warranted about anthropomorphizing LLMs [34], we included a summary prompt with this phrase to test whether it would also facilitate generalization accuracy. The third prompt explicitly asked LLMs to avoid inaccuracies in the summaries.

Since LLM responses can be influenced by temperature, a parameter that controls the randomness of generated text (higher temperatures produce more varied and less constrained outputs), we accessed some models via an API, as this allows explicit temperature control. To maximize replicability and consistency, we retrieved 400 LLM-generated abstract summaries using a temperature setting of 0, the most deterministic setting [35]. However, ChatGPT, the UI for GPT models, is widely assumed to default to a temperature of 0.7, though OpenAI has not disclosed exact details [36,37]. Similarly, the DeepSeek AI Assistant UI does not disclose its default temperature setting (though its API documentation lists 1.0 as the default) [38]. To capture LLM responses as experienced by lay users who do not know how to code (and thus may rely only on the UI), we collected most LLM abstract and article summaries at a temperature of 0.7 or via UIs.

Finally, to assess whether LLM responses remain stable upon retesting, several models were tested multiple times with the same inputs. The details of all conditions (i.e. prompts, temperatures, and retests) and LLM summary retrievals are presented in figure 1, showing that a total of 4900 LLM summaries – 4300 abstract summaries and 600 article summaries – were tested. This total was prespecified to keep data labelling tractable. For the testing, our three main research questions were:

(1) Do algorithmic overgeneralizations occur?
(2) If so, can LLM prompts that focus on systematic ('step-by-step') or accurate processing mitigate them?
(3) Do LLMs differ from human science communicators (specifically, *NEJM JW* authors) in their tendency to overgeneralize?

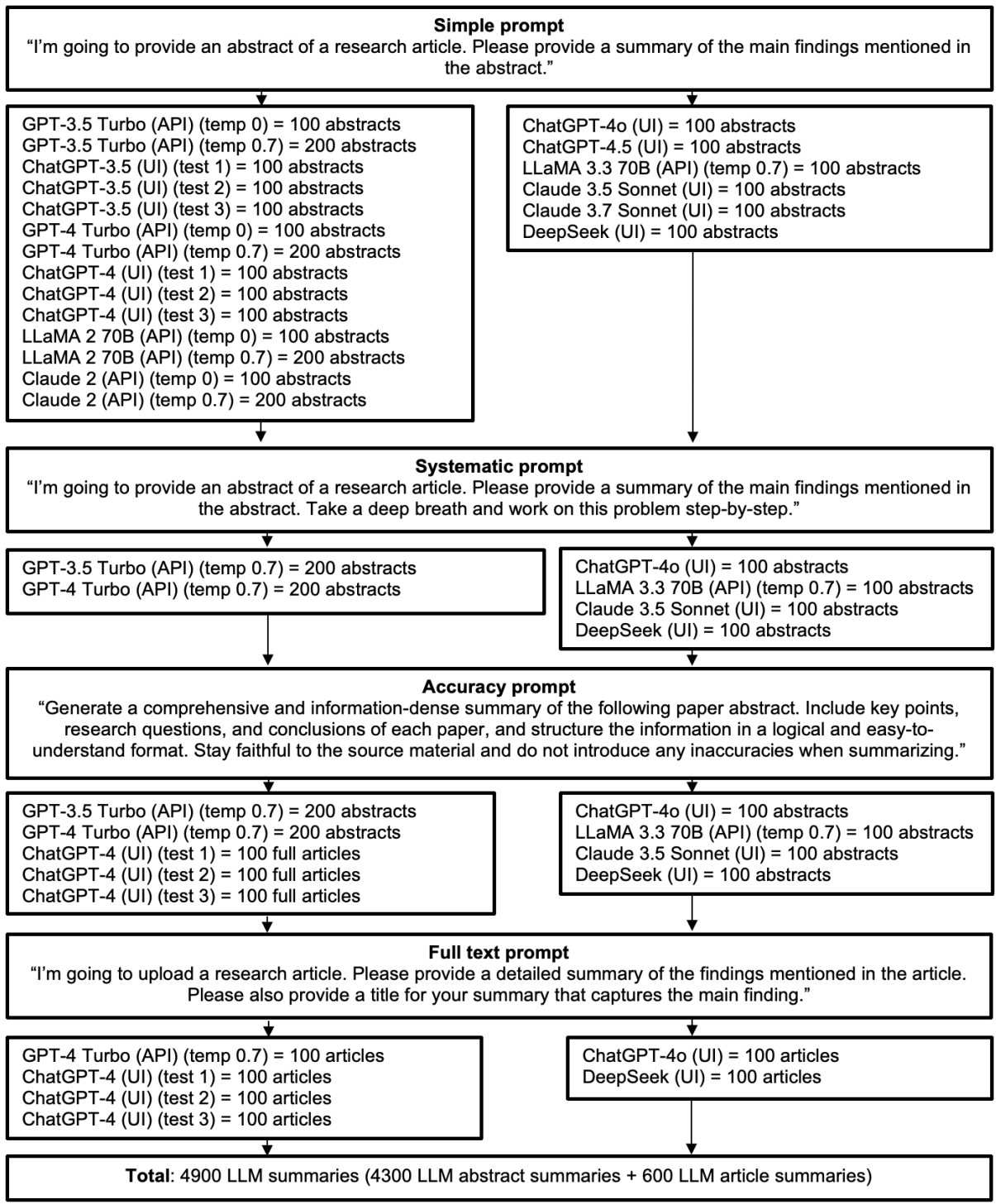

**Figure 1.** Overview of the number of summaries retrieved per LLM, access mode (API, UI), prompt, and temperature setting.

## 2. Results

(1) *Do algorithmic overgeneralizations occur?*

We first compared scientific abstracts and LLM summaries in terms of their likelihood of containing generalized conclusions, combining all original abstracts and their 4300 LLM-generated summaries. A regression analysis was conducted with scope of conclusion (generalized vs. restricted) as the dependent variable and text source (original abstract vs. LLM (all models combined)) as the main predictor, while controlling for temperature, prompt, and test condition (i.e. first test, second test, etc.). The model was significant overall (F7, 4492 = 32.34, p < 0.001), showing that LLM summaries (all combined) were twice as likely to contain generalized conclusions compared to the original abstracts, indicating an algorithmic overgeneralization tendency (table 1, figure 2).

| GLMM regression table | | | | |
|---|---|---|---|---|
| Type of comparison | *B* | *SE* | *t* | *p* |
| **Overall source comparison** | | | | |
| All scientific abstracts as reference | | | | |
| All LLM summaries combined | .693 | .1926 | 3.597 | <.001 |
| **Subtype source comparisons** | | | | |
| GPT-3.5 Turbo (API and UI) vs. original abstracts | .516 | .2719 | 1.896 | .058 |
| GTP-4 Turbo (API and UI) vs. original abstracts | .949 | .2520 | 3.765 | <.001 |
| ChatGPT-4o (UI) vs. original abstracts | 2.200 | .4415 | 4.983 | <.001 |
| ChatGPT-4.5 (UI) vs. original abstracts | .883 | .4438 | 1.989 | .047 |
| LLaMA 2 70B (API) vs. original abstracts | .964 | .2744 | 3.515 | <.001 |
| LlaMA 3.3 70B (API) vs. original abstracts | 3.672 | .3936 | 9.330 | <.001 |
| Claude 2 (API) vs. original abstracts | -.110 | .2707 | -.406 | .685 |
| Claude 3.5 Sonnet (UI) vs. original abstracts | .248 | .4562 | .543 | .587 |
| Claude 3.7 Sonnet (UI) vs. original abstracts | .824 | .4447 | 1.853 | .064 |
| DeepSeek (UI) vs. original abstracts | 1.168 | .4407 | 2.651 | .008 |
| **Temperature comparisons** | | | | |
| Temperature 0.7 as reference | | | | |
| Temp 0.0 vs. temp 0.7 | -1.432 | .3726 | -3.843 | <.001 |
| UI temp vs. temp 0.7 | -.262 | .3219 | -.813 | .416 |
| **Retesting** | | | | |
| Test 1 as reference | | | | |
| Test 2 vs. test 1 | -.199 | .3642 | -.546 | .585 |
| Test 3 vs. test 1 | .426 | .3497 | 1.217 | .224 |
| **Prompt comparisons** | | | | |
| Simple prompt as reference | | | | |
| Systematic vs. simple prompt | -.148 | .2720 | -.544 | .587 |
| Accuracy vs. simple prompt | .640 | .2753 | 2.323 | .020 |
| | | | | |
| **Human versus LLM article summary** | | | | |
| *100 scientific articles as reference* | | | | |
| *NEJM JW* vs. scientific articles | .297 | .2917 | 1.018 | .309 |
| LLMs vs. scientific articles | 1.905 | .2374 | 8.025 | <.001 |
| GPT-4 Turbo (API) (temp 0.7) vs. scientific articles | 1.045 | .3081 | 3.392 | <.001 |
| ChatGPT-4 (UI) test 1 vs. scientific articles | 1.565 | .3306 | 4.735 | <.001 |
| ChatGPT-4 (UI) test 2 vs. scientific articles | 1.501 | .3271 | 4.587 | <.001 |
| ChatGPT-4 (UI) test 3 vs. scientific articles | 2.199 | .3768 | 5.834 | <.001 |
| ChatGPT-4o (UI) vs. scientific articles | 3.176 | .5084 | 6.246 | <.001 |
| DeepSeek (UI) vs. scientific articles | 3.715 | .6259 | 5.934 | <.001 |
| *NEJM JW* summaries as reference | | | | |
| LLMs vs. *NEJM JW* | 1.579 | .2353 | 6.713 | <.001 |
| GPT-4 Turbo (API) (temp 0.7) vs. *NEJM JW* | .728 | .3054 | 2.385 | .017 |
| ChatGPT-4 (UI) test 1 vs. *NEJM JW* | 1.240 | .3278 | 3.781 | <.001 |
| ChatGPT-4 (UI) test 2 vs. *NEJM JW* | 1.176 | .3244 | 3.625 | <.001 |
| ChatGPT-4 (UI) test 3 vs. *NEJM JW* | 1.865 | .3742 | 4.984 | <.001 |
| ChatGPT-4o (UI) vs. *NEJM JW* | 2.835 | .5062 | 5.600 | <.001 |
| DeepSeek (UI) vs. *NEJM JW* | 3.371 | .6241 | 5.402 | <.001 |

**Table 1.** Fixed effects of generalized linear mixed models (GLMMs) predicting the likelihood of generalized (vs. restricted) conclusions in LLM-generated summaries of abstracts and articles. Models comparing abstracts vs. LLM-generated summaries control for temperature, test condition, and prompt type. B coefficients represent unstandardized estimates of each predictor's effect on the likelihood of generalized conclusions, holding other factors constant.

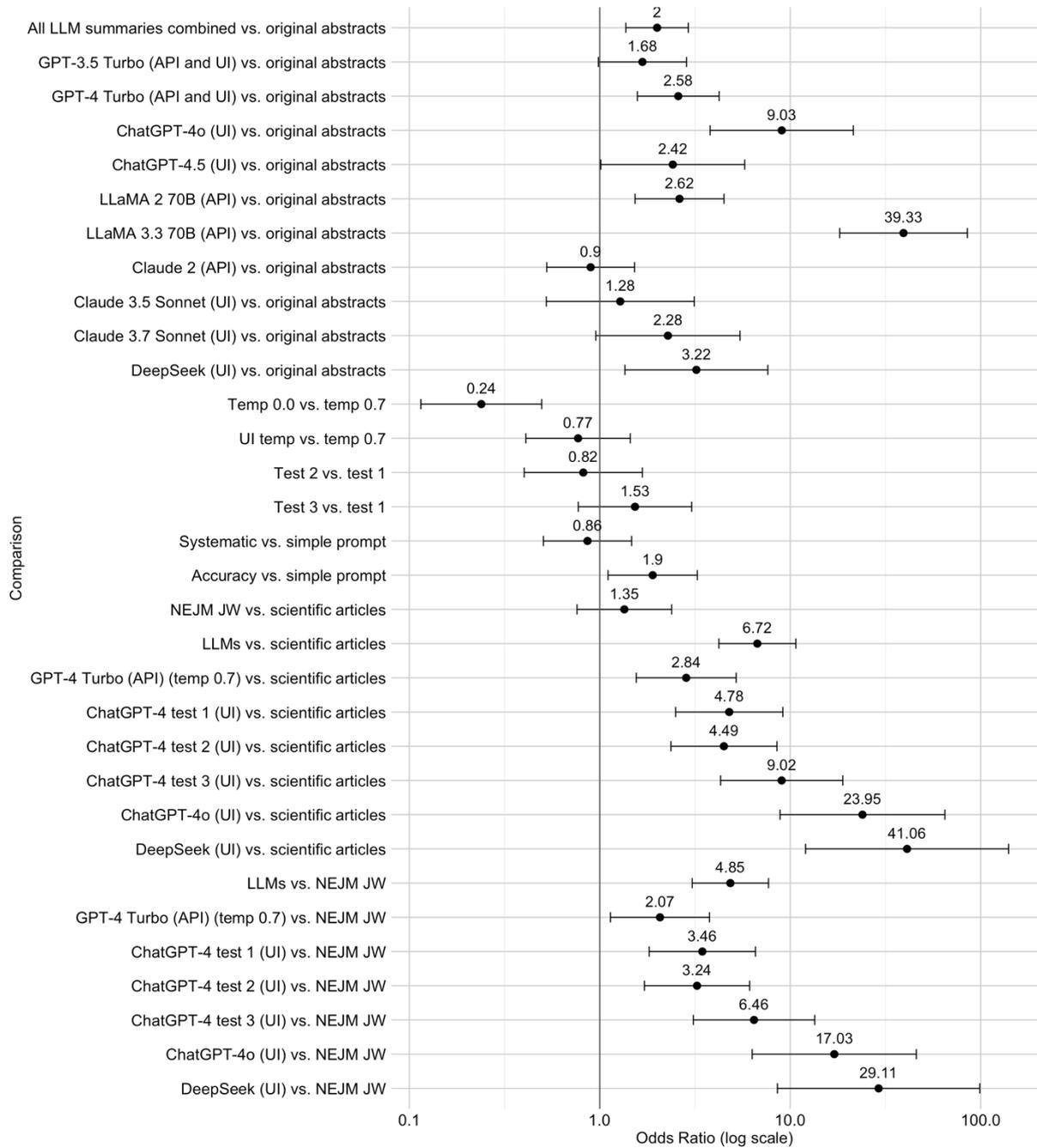

**Figure 2.** Forest plot (based on Table 1) displaying odds ratios (OR) and their 95% confidence intervals for comparisons between LLM-generated summaries, original texts, and human-written summaries (NEJM JW). The plot shows the likelihood of generalized (vs. restricted) conclusions in LLM summaries compared to the corresponding reference texts. Higher ORs reflect stronger overgeneralization tendency. The vertical line at OR = 1 represents no difference from the reference text, indicating the benchmark for fully faithful LLM summaries. Comparisons where error bars overlap this line are not statistically significant.

A subsequent analysis using subtype of text source (original abstract vs. each individual LLM) as the main predictor and controlling for temperature, prompt, test condition, and interactions between LLMs and these three factors showed significant effects of subtype of text source ($F(9, 4467) = 39.58$, $p < 0.001$), temperature ($F(2, 4467) = 4.21$, $p = 0.015$), and prompt ($F(2, 4467) = 17.10$, $p < 0.001$). But there was no evidence that test condition significantly affected LLM generalizations ($F(2, 4467) = 0.56$, $p = 0.57$), suggesting that for the relevant models, the overall results were stable upon retesting.

Focusing on specific models, the summaries by 6 of the 10 models, i.e. GPT-4 Turbo (API and UI), ChatGPT-4o (UI), ChatGPT-4.5 (UI), LLaMA 2 70B (API), LLaMA 3.3 70B (API), and DeepSeek (UI), were significantly more likely to contain generalized conclusions compared to the original texts (table 1, figure 2). From the older models, GPT-4 Turbo (API and UI) and LLaMA 2 70B (API) abstract summaries were about 2.6 times more likely to contain such conclusions compared to the abstracts (figure 2). This tendency increased to 9 (ChatGPT-4o (UI)) and 39 times (LLaMA 3.3 70B (API)) in more recent models. We return to this pronounced difference between older and newer models below. Notably, the summaries by GPT-3.5 Turbo (API and UI) and both the older and most recent versions of Claude (i.e. 2, 3.5 and 3.7) did not significantly differ in generalizations from the abstracts.

Moreover, at LLM temperature 0, summaries containing generalized conclusions were 76% less likely to occur compared to those generated at temperature 0.7 (figure 2). No significant difference was observed between the (unknown) temperature setting of the LLM UIs and temperature 0.7.

Corresponding to the differences in likelihood of producing outputs with generalized conclusions, the number of overall algorithmic overgeneralizations (i.e. the total number of LLM summaries with generalized conclusions higher than the original texts with them) also differed between models (table 2). Newer models such as ChatGPT-4o (UI) (45−60%), LLaMA 3.3 70B (API) (69−73%), and DeepSeek (UI) (26−67%) were associated with the highest proportion of these overgeneralizations, compared to older ones. Claude models had the lowest (−1 to 20%).

Turning to specific algorithmic overgeneralizations (i.e. instances where a specific LLM summary introduced a generic, present tense, or action guiding generalization absent in the original text), table 3 presents concrete examples. Compared to older models (table 4), ChatGPT-4o (UI) and LLaMA 3.3 70B (API) had the highest proportion of specific algorithmic overgeneralizations (reaching 61 and 73%, respectively) (table 5). Claude had consistently the lowest. Notably, across LLMs and prompts, among the tested models, the most frequent transitions from a narrow generalization in the original text to a broader generalization in the LLM summary were transitions from quantified generalizations to generics (table 4).

| Text source | Texts with generalized conclusion | OAO |
|---|---|---|
| **All 200 abstracts** | 108 (54%) | |
| GPT-3.5 Turbo (API) (temp 0.7) | | |
| Simple prompt | 122 (61%) | 14 (7%) |
| Systematic prompt | 118 (59%) | 10 (5%) |
| Accuracy prompt | 139 (69.5%) | 31 (15.5%) |
| GPT-4 Turbo (API) (temp 0.7) | | |
| Simple prompt | 141 (70.5%) | 33 (15.5%) |
| Systematic prompt | 133 (66.5%) | 25 (12.5%) |
| Accuracy prompt | 151 (75.5%) | 43 (21.5%) |
| LLaMA 2 70B (API) (temp 0.7) | | |
| Simple prompt | 134 (67%) | 26 (13%) |
| Claude 2 (API) (temp 0.7) | | |
| Simple prompt | 105 (52.5%) | 0 |
| | | |
| **100 medical abstracts** | 20 (20%) | |
| GPT-3.5 Turbo (API) (temp 0.7) | | |
| Simple prompt | 34 (34%) | 14 (14%) |
| Systematic prompt | 37 (37%) | 17 (17%) |
| Accuracy prompt | 46 (46%) | 26 (26%) |
| GPT-4 Turbo (API) (temp 0.7) | | |
| Simple prompt | 50 (50%) | 30 (30%) |
| Systematic prompt | 50 (50%) | 30 (30%) |
| Accuracy prompt | 56 (56%) | 36 (36%) |
| ChatGPT-4o (UI) | | |
| Simple prompt | 65 (65%) | 45 (45%) |
| Systematic prompt | 75 (75%) | 55 (55%) |
| Accuracy prompt | 80 (80%) | 60 (60%) |
| ChatGPT-4.5 (UI) | | |
| Simple prompt | 41 (41%) | 21 (21%) |
| LLaMA 2 70B (API) (temp 0.7) | | |
| Simple prompt | 51 (51%) | 31 (31%) |
| LlaMA 3.3 70B (API) (temp 0.7) | | |
| Simple prompt | 89 (89%) | 69 (69%) |
| Systematic prompt | 76 (76%) | 56 (56%) |
| Accuracy prompt | 93 (93%) | 73 (73%) |
| Claude 2 (API) (temp 0.7) | | |
| Simple prompt | 19 (19%) | -1 (-1%) |
| Claude 3.5 Sonnet (UI) | | |
| Simple prompt | 31 (31%) | 11 (11%) |
| Systematic prompt | 39 (39%) | 19 (19%) |
| Accuracy prompt | 24 (24%) | 4 (4%) |
| Claude 3.7 Sonnet (UI) | | |
| Simple prompt | 40 (40%) | 20.0 (20.0%) |
| DeepSeek (UI) | | |
| Simple prompt | 46 (46%) | 26.0 (26.0%) |
| Systematic prompt | 68 (68%) | 48.0 (48.0%) |
| Accuracy prompt | 87 (87%) | 67.0 (67.0%) |

**Table 2.** Counts of texts containing generalized conclusions, and overall algorithmic overgeneralizations (OAO)

| **Examples of specific algorithmic overgeneralizations** |
|---|
| *Non-generic to generic generalizations* |
| Original (153): "While exposure to disinformation **had strong detrimental effects on participants' climate change beliefs** ($\delta = -0.16$), affect towards climate mitigation action ($\delta = -0.33$), ability to detect disinformation ($\delta = -0.14$) and pro-environmental behaviour ($\delta = -0.24$), we found almost no evidence for protective effects of the inoculations (all $\delta < 0.20$)." |
| ChatGPT-4 (UI): "The main findings from the experiments indicate that exposure to climate disinformation **significantly undermines individuals' beliefs in climate change**, their positive feelings towards climate mitigation, their ability to recognize disinformation, and their engagement in pro-environmental behaviors." |
| Original (13): "Among adults with obesity, **bariatric surgery** compared with no surgery **was associated** with a significantly lower incidence of obesity-associated cancer and cancer-related mortality." |
| DeepSeek (UI) "The study concluded that **bariatric surgery is associated** with a significantly lower incidence of obesity-associated cancers and cancer-related mortality compared to nonsurgical care in adults with obesity." |
| Original (26): "Among patients undergoing hip or knee arthroplasty for osteoarthritis, **aspirin compared with enoxaparin resulted in a significantly higher rate of symptomatic VTE within 90 days**, defined as below- or above-knee DVT or pulmonary embolism." |
| LLaMA 3.3 70B (API): "Overall, the study suggests that **enoxaparin is more effective than aspirin in preventing symptomatic VTE after THA or TKA.**" |
| |
| *Past to present tense generalizations* |
| Original (7): "Among women with a history of gestational diabetes mellitus, each additional optimal modifiable factor **was associated with an incrementally lower risk of type 2 diabetes**." |
| ChatGPT-4o (UI): "For women with a history of gestational diabetes, every additional healthy lifestyle choice **significantly lowers the risk of type 2 diabetes**, regardless of weight status or genetic predisposition." |
| Original (77): "The consumption of caffeinated coffee **did not result in significantly more** daily premature atrial contractions than the avoidance of caffeine." |
| ChatGPT-3.5 (UI): "The results suggest that the consumption of caffeinated coffee **does not significantly increase** the occurrence of premature atrial contractions." |
| Original (20): "Genital HSV-1 shedding **was frequent** after first-episode genital HSV-1, particularly among those with primary infection, and declined rapidly during the first year after infection" |
| Claude 3.5 Sonnet (UI): "The key takeaway is that genital HSV-1 shedding **is common** initially after first-episode infection, particularly in those with primary infection, but declines significantly during the first year, despite maintained immune responses." |
| |
| *Descriptive to action guiding generalizations* |
| Original (41): "Our results are relevant to clinical practice, **supporting the use of the Mediterranean diet** in secondary prevention." |
| Claude 3.5 Sonnet (UI): "The findings suggest that the **Mediterranean diet should be considered** as a preferred dietary approach for patients with established cardiovascular disease, particularly for men." |
| Original (158): "These findings **have immediate implications for government communicators** and open the door for a renewed focus on how the design and presentation of information impacts behaviour." |
| Claude 2 (API): "The findings suggest **government communicators should consider** using more formal communication designs and presentations, as this can positively impact resident behavior." |
| Original (35): "We found that transcatheter arterialization of the deep veins **was safe and could be performed successfully** in patients with chronic limb-threatening ischemia and no conventional surgical or endovascular revascularization treatment options." |
| DeepSeek (UI): "The study concluded that TADV **is a safe and effective treatment option** for patients with CLTI who lack conventional revascularization options, offering significant benefits in amputation-free survival and limb salvage." |

**Table 3.** Randomly selected examples of transitions from narrower claims in abstracts to generalized conclusions in LLM summaries (simple prompt responses). Numbers in brackets indicate OSF IDs (i.e. identifiers of the text in the datasheets available on our OSF platform). Phrases marking generalization transitions are highlighted in bold.

| Generalizations in LLM summaries of 200 scientific abstracts – Older models | | | | |
|---|---|---|---|---|
| *Prompt and generalization type* | GPT-3.5 Turbo (API, temp 0.7) | GPT-4 Turbo (API, temp 0.7) | LLaMA 2 70B (API, temp 0.7) | Claude 2 (API, temp 0.7) |
| **Simple prompt** | | | | |
| SA overgeneralizations | 23 (11.5%) | 41 (20.5%) | 43 (21.5%) | 10 (5%) |
| SA undergeneralizations | 9 (4.5%) | 8 (4%) | 17 (8.5%) | 13 (6.5%) |
| *Breakdown by generalization type* | | | | |
| (1)* Non-generic to generic | 31 (15.5%) | 51 (25.5%) | 53 (26.5%) | 13 (6.5%) |
| (1)** Generic to non-generic | 13 (6.5%) | 17 (8.5%) | 20 (10%) | 12 (6%) |
| (2)* Past to present tense | 23 (11.5%) | 41 (20.5%) | 43 (21.5%) | 10 (5%) |
| (2)** Present to past tense | 9 (4.5%) | 8 (4%) | 17 (8.5%) | 13 (6.5%) |
| (3)** Descriptive to action guiding | 3 (1.5%) | 5 (2.5%) | 5 (2.5%) | 6 (3%) |
| (3)* Action guiding to descriptive | 4 (2%) | 7 (3.5%) | 10 (5%) | 9 (4.5%) |
| **Systematic prompt** | | | | |
| SA overgeneralizations | 23 (11.5%) | 39 (19.5%) | | |
| SA undergeneralizations | 13 (6.5%) | 14 (7%) | | |
| *Breakdown by generalization type* | | | | |
| (1)* Non-generic to generic | 31 (15.5%) | 41 (20.5%) | | |
| (1)** Generic to non-generic | 15 (7.5%) | 11 (5.5%) | | |
| (2)* Past to present tense | 23 (11.5%) | 39 (19.5%) | | |
| (2)** Present to past tense | 13 (6.5%) | 14 (7%) | | |
| (3)* Descriptive to action guiding | 3 (1.5%) | 1 (0.5%) | | |
| (3)** Action guiding to descriptive | 5 (2.5%) | 10 (5%) | | |
| **Accuracy prompt** | | | | |
| SA overgeneralizations | 33 (16.5%) | 47 (23.5%) | | |
| SA undergeneralizations | 2 (1%) | 4 (2%) | | |
| *Breakdown by generalization type* | | | | |
| (1)* Non-generic to generic | 31 (15.5%) | 47 (23.5%) | | |
| (1)** Generic to non-generic | 6 (3%) | 8 (4%) | | |
| (2)* Past to present tense | 33 (16.5%) | 47 (23.5%) | | |
| (2)** Present to past tense | 2 (1%) | 4 (2%) | | |
| (3)* Descriptive to action guiding | 7 (3.5%) | 10 (5%) | | |
| (3)** Action guiding to descriptive | 5 (2.5%) | 8 (4%) | | |

**Table 4.** Overview of specific algorithmic (SA) overgeneralizations and undergeneralizations by generalization types (SA overgeneralization types indicated with *, SA undergeneralization types indicated with **), focusing on older LLMs. Undergeneralizations are the reverse of overgeneralizations, involving LLMs transitions from broader generalizations in the original text to narrower generalizations in the summary.

| Generalizations in LLM summaries of 100 scientific abstracts – Recent models | | | | | | |
|---|---|---|---|---|---|---|
| *Prompt and generalization type* | ChatGPT-4o (UI) | ChatGPT-4.5 (UI) | LLaMA 3.3 70B (API, temp 0.7) | Claude 3.5 Sonnet (UI) | Claude 3.7 Sonnet (UI) | DeepSeek (UI) |
| **Simple prompt** | | | | | | |
| SA overgeneralizations | 48 (48%) | 26 (26%) | 71 (71%) | 14 (14%) | 22 (22%) | 29 (29%) |
| SA undergeneralizations | 3 (3%) | 5 (5%) | 2 (2%) | 3 (3%) | 2 (2%) | 3 (3%) |
| **Systematic prompt** | | | | | | |
| SA overgeneralizations | 58 (58%) | | 57 (57%) | 26 (26%) | | 51 (51%) |
| SA undergeneralizations | 3 (3%) | | 1 (1%) | 7 (7%) | | 3 (3%) |
| **Accuracy prompt** | | | | | | |
| SA overgeneralizations | 61 (61%) | | 73 (73%) | 8 (8%) | | 67 (67%) |
| SA undergeneralizations | 1 (1%) | | 0 (0%) | 4 (4%) | | 0 (0%) |

**Table 5.** Overview of specific algorithmic (SA) overgeneralizations and undergeneralizations, focusing on recent LLMs.

(2) *Can LLM prompts that focus on systematic or accurate processing mitigate algorithmic overgeneralizations?*

Compared to the simple prompt, the systematic prompt did not significantly change the likelihood of LLM outputs containing generalized conclusions. However, the accuracy prompt did change the chances, albeit in an unexpected direction: LLM summaries retrieved with the accuracy prompt were about twice as likely to contain generalized conclusions compared to the simple prompt (OR = 1.90, 95% CI [1.11, 3.26], *p* = 0.02) (figure 2). Correspondingly, for all models (older and newer versions), except Claude, the proportion of both overall and specific algorithmic overgeneralizations was also highest when the accuracy prompt was used (table 4 and table 5).

(3) *Do LLMs differ from human science communicators in producing overgeneralizations?*

Previous studies found that human science communicators also often overgeneralize or exaggerate research results [16–18]. To examine whether LLMs differ from humans in this respect, we additionally tested GPT-4 Turbo (API and UI), ChatGPT-4o (UI), and DeepSeek (UI) on its summarization of 100 full-length scientific (medical) articles that had corresponding human-authored summaries published in *NEJM JW*, enabling direct human-LLM summary comparisons.

Compared to the original articles, *NEJM JW* summaries did not significantly differ in their likelihood of containing generalized conclusions (table 1). However, overall, LLM (GPT-4 Turbo (API and UI), ChatGPT-4o (UI), and DeepSeek (UI)) article summaries had more than 6 times higher chances of containing generalized conclusions than the articles themselves (OR = 6.72, 95% CI [4.22, 10.71], *p* < 0.001). Moreover, when the human-authored *NEJM JW* summaries of the same articles were used as the reference, LLM summaries were almost 5 times as likely to contain generalized conclusions compared to *NEJM JW* summaries (OR = 4.85, 95% CI [3.06, 7.70], *p* < 0.001). This likelihood (figure 2), as well as the number of overall and specific algorithmic overgeneralizations (figure 3), substantially increased in newer models such as ChatGPT-4o and DeepSeek.

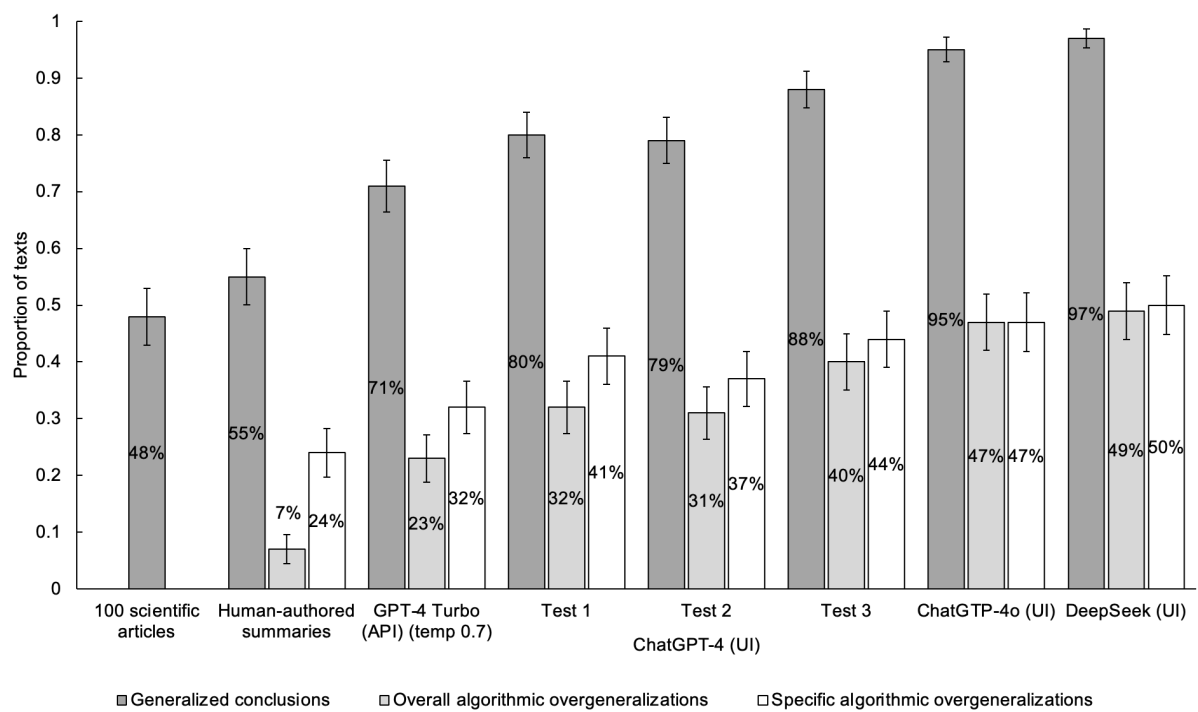

**Figure 3.** Comparisons between the raw proportions of scientific articles and human-authored as well as LLM-generated article summaries that contain generalized conclusions, overall algorithmic overgeneralizations, and specific algorithmic overgeneralizations, presented by text source and test condition. Error bars represent standard errors.

## 3. Discussion

While LLMs hold substantial potential as tools for scientific summarization [3,5], their use carries significant risks, as they may oversimplify or exaggerate scientific findings [12], which can lead to large-scale misunderstandings of science. Until now, this has not been systematically investigated. Our analysis provides the first evidence of these risks, revealing three key findings.

(1) *Algorithmic overgeneralizations occurred frequently and increased in newer models.*

When GPT-4 Turbo (API and UI), ChatGPT-4o (UI), ChatGTP-4.5 (UI), LLaMA models, and DeepSeek (UI) summarized abstracts and GPT-4 Turbo (API and UI), ChatGPT-4o (UI), and DeepSeek (UI) summarized full-length articles, they reliably made broader claims than the original texts. While these claims often contained hedges and subtle scope extensions (e.g. 'suggests', 'may', 'can lead to'), they remain problematic because the authors of the original texts may have refrained from such generalizations due to a lack of evidential support for them.

Moreover, the most common specific algorithmic overgeneralizations were transitions from quantified generalizations to generics. Generics are known to pose special risks in science communication, as they gloss over variations between individuals [17,18], can promote stereotyping [39] and may cause real-world harm [40]. For example, the claim generated by LLaMA 3.3 70B (API) that 'dulaglutide is an effective and safe treatment option for improving glycaemic control in youths with type 2 diabetes' (OSF ID 76) implies much broader efficacy and safety than the original text, which concluded: 'dulaglutide at a once-weekly dose of 0.75 or 1.5 mg was superior to placebo in improving glycaemic control through 26 weeks among youths with type 2 diabetes who were being treated with or without metformin or basal insulin, without an effect on BMI'. Such transitions to generic generalizations could mislead practitioners into using unsafe interventions [26]. That said, this tendency towards overgeneralization was not observed in summaries generated by Claude models, which did not significantly differ from the original texts, a finding consistent across newer and older models. This aligns with prior research reporting that, among popular LLMs, Claude was the most faithful in text summarization [20]. Importantly, the newer LLMs we tested (except ChatGPT-4.5 (UI)) exhibited a stronger tendency to overgeneralize. While counterintuitive (as one might expect LLMs to become more accurate over time), our result aligns with and extends recent findings that earlier LLMs were more likely to avoid answering difficult questions, whereas newer, larger, and more instructible models, instead of refusing to answer, often produced misleadingly authoritative yet flawed responses [41].

The decrease in generalization accuracy we observed may stem from two inter-related factors that can arise when models are fine-tuned for adaptability. First, models may undergo what computer scientists call 'catastrophic forgetting', where learning new information disrupts previously acquired knowledge or skills [42].

Second, as larger LLMs are optimized for helpfulness, they may prioritize generating responses that seem plausible, contextually relevant and widely applicable over strict accuracy. For instance, studies have found that while reinforcement learning from human feedback (RLHF) enhanced models' helpfulness, it often led them to express unwarranted confidence [43] or reduced their ability to hedge claims to indicate uncertainty [44]. Similarly, LLMs frequently failed to refuse or express uncertainty about questions beyond their parametric knowledge [45]. During RLHF fine-tuning, human evaluators may favour responses that are confident and broadly applicable. If a model hedges, expresses uncertainty, or provides a highly precise but complex answer, it may receive lower ratings from human evaluators. Consequently, models may learn to prioritize confident fluency over caution and precision, increasing their tendency to produce overgeneralized statements. This, in turn, could become a strategy for appearing maximally helpful to the broadest range of users.

If this holds true, one might expect LLMs still in development – released primarily to monitor user interactions and refine future versions – to exhibit less overgeneralization than fully developed models. Our results align with this prediction, as the most recent LLM we tested, ChatGPT-4.5 (UI), was only accessible as a 'Research Preview' model (i.e. it is still in development) but showed a reduced tendency to overgeneralize compared to the other new models (which are no longer preview models). These trends in newer, fully developed LLMs raise particular concerns for LLM-based scientific summaries, where readers may struggle to distinguish accurate conclusions from algorithmic overgeneralizations.

(2) *Asking LLMs for faithful science summaries increased algorithmic overgeneralizations.*

Explicitly requesting accurate responses from LLMs seems intuitive to retrieve summaries that capture all relevant details of input texts. However, we found that this backfired. Compared to a simple summarization request, asking for responses faithful to the original text produced a twofold increase in the likelihood of generalized conclusions, in some models, increasing overall algorithmic overgeneralizations by up to 15% (see e.g. ChatGPT-4o (UI), table 2). This finding extends previous research that suggests adding information intended to improve LLM accuracy in LLM prompts can be counterproductive [46].

One potential explanation of this backfire effect may be related to the content of the accuracy prompt, which stated 'do not introduce any inaccuracies when summarizing' (figure 1). Psychological research on humans found 'ironic rebound' effects, which can occur when people attempting to free their mind of a target thought experience higher levels of occurrence and accessibility of the thought compared with people intentionally focusing on the thought (e.g. 'Don't think of a pink elephant now' [47]). While this remains to be tested and caution about LLM anthropomorphism is warranted [34], our prompt may have triggered a similar tendency in LLMs. By calling for attempts to free processing of inaccuracy, it may have made the occurrence of inaccuracy more likely, suggesting an algorithmic version of the 'ironic rebound' effect.

(3) *LLMs' overgeneralization tendency was robust on retesting and, for GPT-4 Turbo (API and UI), ChatGPT-4o (UI), and DeepSeek (UI), more pronounced than in human authors writing summaries for NEJM JW.*

Our finding that most LLMs consistently produced algorithmic overgeneralizations across multiple occasions adds nuance to prior research that found that, in some domains, LLMs produced unstable or inconsistent responses over time [48–50]. These previous inconsistencies concerned the specific content of LLM responses (e.g. moral judgements [50] or reasoning tasks [48]). We instead focused on recurrent linguistic structures in LLM outputs, i.e. generic, present tense, or action guiding conclusions. That most models produced overgeneralizations reliably across time suggests that they may have incorporated an 'algorithmic bias' from the training data similar to the way models trained on linguistic data can acquire gender, racial, or political biases, by picking up human tendencies implicit in training text corpora [51–53].

Relatedly, recent research found that generalizations of study results beyond the evidence provided by a given study were common in scientific articles across different disciplines [16–18]. This pattern has been taken to indicate that many scientists may be vulnerable to a 'generalization bias' [54]. If we treat the conclusions of original texts as the normative baseline for accurate LLM-generated summaries, then our results suggest that an algorithmic version of this bias may also affect LLMs.

In fact, our direct comparison between human-authored *NEJM JW* and LLM summaries of the same articles showed that GPT-4 Turbo (API and UI), ChatGPT-4o (UI), and DeepSeek (UI) (all combined) were significantly more prone to overgeneralizations. Our finding challenges previous studies on clinical text summarization which reported that 'summaries from our best-adapted LLMs (including

GPT-3.5, GPT-4, and LLaMA 2) are preferable to human summaries in terms of completeness and correctness' [19, p. 3].

That said, our comparative results for article summaries pertain only to a specific subset of human-written science summaries, as *NEJM JW* authors are domain experts writing for an expert audience. Their approach may differ from that of other professional science communicators, such as those in university media offices, public relations, or marketing departments. In these contexts, incentives to 'hype' research findings (e.g. to attract attention, funding, or prestige) may be stronger, leading to more frequent overgeneralizations.

Relatedly, LLMs prompted to adopt the role of a domain expert, university media office, or marketing writer may exhibit varying rates of overgeneralization. However, in our study, we did not instruct LLMs to assume any specific role. We welcome future research exploring these differences to improve comparability between human- and LLM-generated summaries of scientific articles.

Nevertheless, since scientific abstracts are paradigmatic human-authored science summaries, and our LLM prompts explicitly requested systematic, detailed, and faithful abstract summaries, one might expect the LLM outputs to closely reflect the original text without significant overgeneralization. Yet, they did not. Our findings thus offer novel contributions to research on overgeneralization in science communication [17,18,54] and the shortcomings of LLMs in text summarization [48,52,55], revealing a subtle form of 'hallucination' that has not yet been documented in the literature.

## 4. Recommendations

Based on our findings, we propose five strategies that LLM users and developers may explore to mitigate the risks of algorithmic overgeneralizations in scientific text summarization.

(1) *Use a conservative temperature.* While many LLM UIs do not allow users to adjust temperature settings, those accessing models via APIs or platforms such as GroqCloud (which enables temperature adjustments without coding experience) may consider setting the temperature to 0 for high-stake tasks [56]. Our findings show that lower temperatures reduced algorithmic overgeneralization.

(2) *Consider potential backfire effects of accuracy prompts.* Since prompts that included direct requests to avoid inaccuracy increased algorithmic overgeneralizations, users may wish to refrain from such phrasing when prompting LLMs for scientific text summaries.

(3) *Use Claude or explore older LLMs for science summarization.* Since Claude's outputs remained closest to the original text in generalization scope, it may be a preferable model for summarizing scientific literature. Additionally, since older models (e.g. GPT-3.5 Turbo (API and UI)) tended to produce summaries more closely aligned with the original texts than newer, larger models (except ChatGTP-4.5 (UI), which is still in development), using older models instead could help mitigate the problematic tendencies discussed.

(4) *Implement prompt transformation for past tense summarization of scientific results.* Prompt transformation, or 'shadow prompting', involves developers programming their models so that the systems alter the prompts that the user types in before a response is being generated [57]. While this approach is often used to ensure that LLM responses align with ethical guidelines [58], if they enforce past tense summaries of scientific studies, 'shadow prompts' could also prevent models from inaccurately extending past findings into the present. Preference for past tense reporting is already common in top medical journals and has also been recommended for human science communicators [18,54].

(5) *Benchmark LLMs using our three-step framework.* Our study introduces a systematic evaluation method for measuring LLM generalization accuracy. This framework involves: (1) prompting an LLM to summarize scientific texts, (2) classifying both the original texts and LLM summaries based on three key features (generic, present tense, and action guiding generalizations) and (3) comparing these classifications to detect instances where LLMs broaden claims beyond their original scope, generating an overgeneralization score (e.g. an OAO score). This quantifiable benchmark may enable developers to test, compare, and refine LLMs to ensure they generate more accurate science summaries.

## 5. Strengths and limitations

Our study has several key strengths, including a large sample of 4900 LLM-generated summaries and an evaluation of 10 influential models across different temperatures and prompts. Additionally, by analyzing both earlier and the latest models (e.g. DeepSeek, ChatGPT-4o, ChatGPT-4.5, and Claude 3.7 Sonnet), this study provides diachronic insights into how LLM performance has evolved over time, suggesting that, for most tested models, the trend appears to be toward, rather than away from, problematic generalizations. Even if our findings are specific to the models tested, as noted, this study also offers a methodological contribution, introducing a conceptual framework that can be applied to any future LLM, helping developers and users systematically test LLMs for overgeneralization tendencies and optimize model performance.

However, we tested only three different prompts, chosen based on their intuitiveness or existing literature [32]. Enhanced prompt testing and engineering is important to identify mitigation strategies.

Additionally, other models than the ones we tested could be used for science summarization. Also, to compare human- versus LLM-generated summaries of full articles (vs. abstracts), we compared only GPT-4 Turbo (API and UI), ChatGPT-4o (UI), and DeepSeek (UI) summaries to *NEJM JW* summaries. As noted, *NEJM JW* may not be a representative of all human science summarization and other LLMs may be more accurate.

Furthermore, while our key analyses with 200 abstracts included 100 abstracts from multidisciplinary science articles, most other analyses focused on 100 medical abstracts and 100 full-length medical articles (figure 1), as accurate clinical research summarization is particularly critical due to its potential influence on policy-making and patient care. However, the tendency of both humans and LLMs to overgeneralize when summarizing scientific texts may vary by subject matter (e.g. foundational versus applied research). Future research should examine how LLM summarization differs across scientific text domains.

Finally, while our study focused on cases where LLMs produce overly broad generalizations, they may also undergeneralize. For example, an LLM might transform a generic, present tense statement in the original text into a quantified, past tense claim, even when the original generalization was warranted by the research. Although generalization errors can occur in both directions, as shown in table 4 and table 5, overgeneralizations were far more frequent, justifying our primary focus on them.

## 6. Conclusion

To our knowledge, this study is the first to systematically evaluate whether prominent LLMs, including ChatGPT, DeepSeek, and Claude, faithfully summarize scientific claims or exaggerate their scope. Our analysis of nearly 5000 LLM-generated science summaries revealed that most models produced broader generalizations of scientific results than the original texts – even when explicitly prompted for accuracy and across multiple tests. Notably, newer models exhibited significantly greater inaccuracies in generalization than earlier versions. These findings suggest a persistent generalization bias in many LLMs, i.e. a tendency to extrapolate scientific results beyond the claims found in the material that the models summarize, underscoring the need for stronger safeguards in AI-driven science summarization to reduce the risk of widespread misunderstandings of scientific research.

## 7. Methods

This experimental study, which was preregistered on an OSF platform here, combined between- and within-subject aspects, testing different and the same LLMs multiple times. The 10 tested LLMs were accessed through either an API or UI (figure 1). In UI data collections, LLM summaries were retrieved in separate chats, either using new accounts or with memory turned off (ChatGPT-4o) to mitigate personalization.

*Material.* 200 abstracts of scientific articles were used: 100 from the top four general medical journals (*Lancet*, *NEJM*, *JAMA*, and the *BMJ*) and 100 from the top four multidisciplinary science journals (*Nature*, *Science*, *Nature Human Behavior*, and *Psychological Science in the Public Interest*) as ranked by the 2022/23 Clarivate Journal Citation Reports. The 25 most recent abstracts from each journal were collected by moving backward from December 2023, excluding non-research articles (e.g. opinion pieces and commentaries).

For more generalizable results, 100 full-length articles were added to test GPT-4 Turbo (API and UI), ChatGPT-4o (UI), and DeepSeek (UI) article summarization. They were taken from the four medical journals (25 per journal, moving back from May 2023), focusing only on original prospective clinical studies, as they offer key evidence for the efficacy of medical interventions, making their summaries particularly relevant. For these 100 studies, corresponding *NEJM JW* summaries were also collected. LLM summaries were retrieved with four prompts and different model temperatures as shown in figure 1.

For retrieving GPT-4 Turbo (API and UI), ChatGPT-4o (UI), and DeepSeek (UI) article summaries, we used a version of the simple prompt designed to ensure comparability with NEJM JW summaries, which also always have a title.

Procedure. After retrieving LLM summaries, two experts in corpus analysis and science communication coded each text as containing either restricted or generalized conclusions using preregistered criteria (see OSF material). A third, independent researcher, blinded to the summary source, applied the same criteria to 100 texts. Inter-rater agreement ranged from k = 0.79, 95% CI [0.70, 0.87] to k = 0.95, 95% CI [0.91, 0.99]. Disagreements were resolved through discussion. All generalized conclusions identified by the researchers were recorded in spreadsheets available [here](#).

*Statistical information.* To analyze the distribution of generalized conclusions, we modelled the probability of a text containing them (categorical dependent variable) using generalized linear mixed models (GLMMs) with a binomial distribution and logit link. To avoid multicollinearity problems and tailor analyses to the different text types (abstract and article summaries), six separate models were conducted with Bonferroni corrections (for models (1) and (2), $\alpha = 0.025$; for models (3) to (6), $\alpha = 0.0125$).

Model (1) compared the probability of generalized conclusions in LLM summaries vs. abstracts, using overall source (abstracts versus all LLM summaries) as the main predictor with temperature (0, 0.7, UI temperature), test condition (tests 1−3) and prompt (simple, systematic, and accuracy) as fixed effects and a unique identifier for each abstract as a random intercept to account for repeated measures. Model (2) used the same variables but divided the LLM source category into individual LLMs to test for differences. Interaction terms (source*temperature, source*test, and source*prompt) were included to determine if the effects of temperature, test, and prompt were different depending on the source type. Model (3) compared 100 articles and their generalized conclusions to the corresponding *NEJM JW*, GPT-4 Turbo (API and UI), ChatGPT-4o (UI), and DeepSeek (UI) summaries, combining all 600 LLM responses to assess overall effects. Main predictor was text source (article, *NEJM JW*, or LLM (all LLMs combined, i.e. GPT-4 Turbo, ChatGPT-4o, and DeepSeek)), with a random effect for each article. Model (4) used the same set-up but with subtype of text source as main predictor, separating the individual LLMs (and LLM tests). Models (5)–(6) repeated this approach, using *NEJM JW* summaries as the reference instead of the articles.

Model assumptions, including independence of observations, linearity of the logit, absence of multicollinearity and random effect significance, were assessed and met. Independence was ensured by including original text identifiers as a random effect, with source, temperature, test condition, and prompt as fixed effects. Linearity of the logit was assumed for the binary outcome (generalized vs. restricted) and confirmed through model fit statistics. Variance inflation factors were within acceptable limits, indicating no multicollinearity among the fixed effects. The random effect was significant, supporting its inclusion. Analyses and visualizations were done using IBM SPSS 29.0 and R Studio

## Conflicting Interest Statement

The authors have no conflicting interests to declare.

## Data availability

All data, LLM responses, etc. are available on an OSF platform <u>here</u>.

## Author Contributions

U.P. is the main author, conceptualized and designed the study, collected and analyzed the data, developed the main interpretations and arguments, prepared the first draft, and revised manuscript. B.C.-Y. contributed to conceptualization and design of the study, collected the data, and revised the manuscript.

## Acknowledgments

Many thanks to Ushnish Sengupta, Olivier Vroome, and Andrea Bertalozzi for their assistance with the LLM data collection, Olivier Lemeire for assistance with the data labelling, and Oliver Braganza and Chiara Lisciandra for feedback on the project.